\documentclass[11pt,a4paper]{article}
\usepackage[utf8]{inputenc}
\usepackage[hyphens]{url}
\usepackage[]{hyperref}
\usepackage{acl2017}
\usepackage{times}
\usepackage{latexsym}
\usepackage{graphicx}
\usepackage{url}
\usepackage{amsmath}
\usepackage{color}
\usepackage[linesnumbered,algoruled,boxed,lined]{algorithm2e}
\usepackage{float}
\usepackage{framed}

\aclfinalcopy 

\title{Fortia-FBK at SemEval-2017 Task 5:\\Bullish or Bearish? \\ Inferring Sentiment towards Brands from Financial News Headlines}

\author{Youness Mansar$^\star$, Lorenzo Gatti$^\circ$, Sira Ferradans$^\star$, Marco Guerini$^\circ$, and Jacopo Staiano$^\star$\\
$^\star$Fortia Financial Solutions, Paris, France\\ 
$^\circ$Fondazione Bruno Kessler, Povo, Italy\\
              \{youness.mansar,sira.ferradans,jacopo.staiano\}@fortia.fr\\ \{l.gatti,guerini\}@fbk.eu\\
               }

\begin{document}

\maketitle

\begin{abstract}
    In this paper, we describe a methodology to infer \emph{Bullish} or \emph{Bearish} sentiment towards companies/brands. More specifically, our approach leverages affective lexica and word embeddings in combination with convolutional neural networks to infer the sentiment of financial news headlines towards a target company.
    Such architecture was used and evaluated in the context of the SemEval 2017 challenge (task 5, subtask 2), in which it obtained the best performance.
\end{abstract}

\section{Introduction}
Real time information is key for decision making in highly technical domains such as finance.
The explosive growth of financial technology industry (\emph{Fintech}) continued in 2016, partially due to the current interest in the market for Artificial Intelligence-based technologies\footnote{F. Desai, ``The Age of Artificial Intelligence in Fintech" \url{https://www.forbes.com/sites/falgunidesai/2016/06/30/the-age-of-artificial-intelligence-in-fintech}

S. Delventhal, ``Global Fintech Investment Hits Record High in 2016" \url{http://www.investopedia.com/articles/markets/061316/global-fintech-investment-hits-record-high-2016.asp}}.

Opinion-rich texts such as micro-blogging and news can have an important impact in the financial sector (\emph{e.g.} raise or fall in stock value) or in the overall economy (\emph{e.g.} the Greek public debt crisis). In such a context, having granular access to the opinions of an important part of the population is of key importance to any public and private actor in the field. 
In order to take advantage of this raw data, it is thus needed to develop machine learning methods allowing to convert unstructured text into information that can be managed and exploited. 

In this paper, we address the sentiment analysis problem applied to financial headlines, where the goal is, for a given news headline and target company, to infer its polarity score \emph{i.e.} how positive (or negative) the sentence is with respect to the target company. Previous research~\cite{goonatilake2007volatility} has highlighted the association between news items and market fluctiations; hence, in the financial domain, sentiment analysis can be used as a proxy for \emph{bullish} (i.e. positive, upwards trend) or \emph{bearish} (i.e. negative, downwards trend) attitude towards a specific financial actor, allowing to identify and monitor in real-time the sentiment associated with e.g. stocks or brands. 

Our contribution leverages pre-trained word embeddings (\texttt{GloVe}, trained on wikipedia+gigaword corpus), the \texttt{DepecheMood} affective lexicon, and convolutional neural networks. 

\section{Related Works}
While image and sound come with
a natural high dimensional embedding, the issue of \emph{which is the best representation} is still an open research problem in the context of natural language and text. It is beyond the scope of this paper to do a thorough overview of word representations, for this we refer the interest reader to the excellent review provided by~\cite{mandelbaum2016word}. Here, we will just introduce the main representations that are related to the proposed method. 

\paragraph{Word embeddings.}In the seminal paper~\cite{bengio2003neural}, the authors introduce a statistical language model computed in an unsupervised training context using shallow neural networks. 
The goal was to predict the following word, given the previous context in the sentence, showing a major advance with respect to n-grams. Collobert \emph{et al.}~\cite{collobert2011natural} empirically proved the usefulness of using unsupervised word representations for a variety of different NLP tasks and set the neural network architecture for many current approaches. Mikolov \emph{et al.}~\cite{mikolov2013efficient} proposed a simplified model (\texttt{word2vec}) that allows to train on larger corpora, and showed how semantic relationships emerge from this training. Pennington \emph{et al.}~\cite{pennington2014glove}, with the \texttt{GloVe} approach, maintain the semantic capacity of \texttt{word2vec} while introducing the statistical information from latent semantic analysis (LSA) showing that they can improve in semantic and syntactic tasks.

\paragraph{Sentiment and Affective Lexica.}
In recent years, several approaches have been proposed to build lexica containing prior 
sentiment polarities (sentiment lexica) or multi-dimensional affective scores (affective lexica). The goal of these methods is to associate such scores to raw tokens or tuples, e.g. \texttt{lemma\#pos} where \texttt{lemma} is the lemma of a token, and \texttt{pos} its part of speech. 

There is usually a trade-off between coverage (the amount of entries) and precision (the accuracy of the sentiment information). For instance, regarding sentiment lexica, \emph{SentiWordNet} ~\cite{Esuli06},~\cite{baccianella2010sentiwordnet}, associates each entry with the numerical scores, ranging from 0 (negative) to 1 (positive); following this approach, it has been possible to automatically obtain a list of 155k words, compensating a low precision with a high coverage~\cite{gatti2016sentiwords}. 
On the other side  of the spectrum, we have methods such as  ~\cite{bradley1999affective},~\cite{taboada2011lexicon},~\cite{warriner2013norms} with low coverage (from 1k to 14k words), but for which the precision is maximized. These scores were manually assigned by multiple annotators, and in some cases validated by crowd-sourcing~\cite{taboada2011lexicon}. 



Finally, a binary sentiment score is provided in the \emph{General
Inquirer} lexicon~\cite{stone1966general}, covering 4k
sentiment-bearing words, and expanded to 6k words by~\cite{wilson2005recognizing}. 

Turning to affective lexica, where multiple dimensions of affect are taken into account, we mention \emph{WordNetAffect}~\cite{strappaLREC04}, which provides manual affective annotations of WordNet synsets 
(\textsc{anger}, \textsc{joy}, \textsc{fear}, etc.): it contains 900 annotated synsets and 1.6k words in the form \texttt{lemma\#PoS\#sense}, which correspond to roughly 1k \texttt{lemma\#PoS} entries. 

\emph{AffectNet}~\cite{cambria2012sentic}, contains 10k words taken from ConceptNet and aligned with WordNetAffect, and extends the latter to concepts like `have breakfast'.
\emph{Fuzzy Affect Lexicon} \cite{subasic2001affect} contains roughly 4k \texttt{lemma\#PoS} manually annotated by one linguist using 80 emotion labels.
\emph{EmoLex} \cite{mohammad2013crowdsourcing} contains almost 10k lemmas annotated with an intensity label for each emotion using Mechanical Turk.
Finally, \emph{Affect database} is an extension of \emph{SentiFul}~\cite{Neviarouskaya:2007fk} and contains 2.5k words in the form \texttt{lemma\#PoS}. The latter is the only lexicon 
providing words annotated also with emotion scores rather than only with labels.

In this work, we exploit the \texttt{DepecheMood} affective lexicon proposed by~\cite{staiano2014depeche}: this resource has been built in a completely unsupervised fashion, from affective scores assigned by readers to news articles; notably, due to its automated crowd-sourcing-based approach, \texttt{DepecheMood} allows for both high-coverage and high-precision. \texttt{DepecheMood} provides scores for more than 37k entries, on the following affective dimensions: \emph{Afraid}, \emph{Happy}, \emph{Angry}, \emph{Sad}, \emph{Inspired}, \emph{Don't Care}, \emph{Inspired}, \emph{Amused}, \emph{Annoyed}. We refer the reader to~\cite{staiano2014depeche,Guerini2015_www} for more details.

The affective dimensions encoded in \texttt{DepecheMood} are directly connected to the emotions evoked by a news article in the readers, hence it seemed a natural choice for the SemEval 2017 task at hand. 

\paragraph{Sentence Classification.} 
A modification of~\cite{collobert2011natural} was proposed by Kim~\cite{kim2014convolutional} for sentence classification, showing how a simple model together with pre-trained word representations can be highly performing. Our method builds on this conv-net method. Further, we took advantage of the 
rule-based sentiment analyser \texttt{VADER}~\cite{Vader} (for Valence Aware Dictionary for
sEntiment Reasoning), which builds upon a sentiment lexicon and a predefined set of simple rules.

\section{Data}
The data consists of a set of financial news headlines, crawled from several online outlets such as Yahoo Finance, where each sentence contains one or more company names/brands. 

Each tuple (headline, company) is annotated with a sentiment score ranging from -1 (very negative, bearish) to 1 (very positive, bullish). The training/test sets provided contain 1142 and 491 annotated sentences, respectively. 

A sample instance is reported below:
\begin{framed}
\begin{quote}
Headline: \emph{``Morrisons book second consecutive quarter of sales growth''} 

Company name: \emph{``Morrisons''}

Sentiment score: \emph{0.43}
\end{quote}
\end{framed}

\section{Method}
\begin{figure}[h!]
  \centering
    \includegraphics[width=0.33\textwidth]{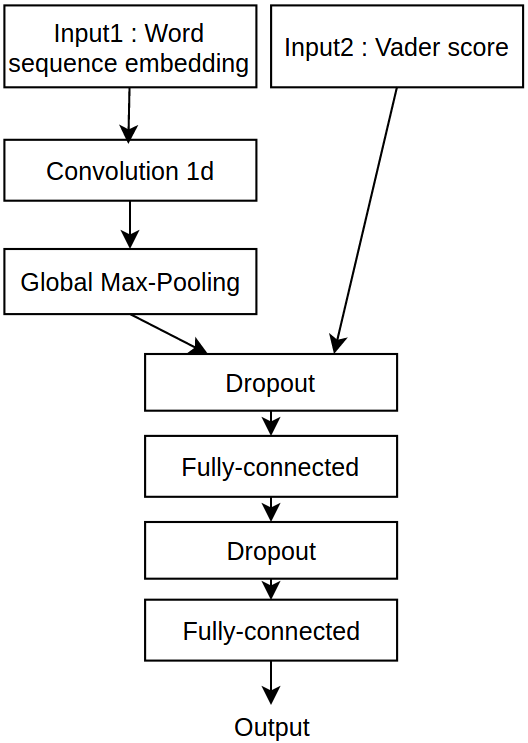}
    \caption{Network architecture}
    \label{fig:model}
\end{figure}

In Figure~\ref{fig:model}, we can see the overall architecture of our model. 

\subsection{Sentence representation and preprocessing}
\paragraph{Pre-processing.} Minimal preprocessing was adopted in our approach: we replaced the target company's name with a fixed word \texttt{\textless company\textgreater} and numbers with \texttt{\textless number\textgreater}. The sentences were then tokenized using spaces as separator and keeping punctuation symbols as separate tokens.

\paragraph{Sentence representation.} The words are represented as fixed length vectors $u_i$ resulting from the concatenation of \texttt{GloVe} pre-trained embeddings and \texttt{DepecheMood}~\cite{staiano2014depeche} lexicon representation. Since we cannot directly concatenate token-based embeddings (provided in \texttt{GloVe}) with the \texttt{lemma\#PoS}-based representation available in \texttt{DepecheMood}, we proceeded to re-build the latter in token-based form, applying the exact same methodology albeit with two differences: we started from a larger dataset (51.9K news articles instead of 25.3K) 
 and used a frequency cut-off, \emph{i.e.} keeping only those tokens that appear at least 5 times in the corpus\footnote{Our tests showed that: (i) the larger dataset allowed improving both  precision on the SemEval2007 Affective Text Task~\cite{strapparava2007semeval} dataset, originally used for the evaluation of \texttt{DepecheMood}, and coverage (from the initial 183K unique tokens we went to 292K entries) of the lexicon; (ii) we found no significant difference in performance between \texttt{lemma\#PoS} and token versions built starting from the same dataset.}. 

These word-level representation are used as the first layer of our network. During training we allow the weights of the representation to be updated. We further add the \texttt{VADER} score for the sentence under analysis. The complete sentence representation is presented in Algorithm~\ref{algo:repr}.

\begin{algorithm}
\SetAlgoLined
    \SetKwInOut{Input}{Input}
    \SetKwInOut{Output}{Output}
    \Input{An input sentence $s$, and the GloVe word embeddings $W$}
    \Output{The sentence embedding $x$}
    $v = {\rm VADER}(s)$\\
    \ForEach{$w_i$ in $W$}{$u_i$ = [GloVe($w_i$,$W$), DepecheMood($w_i$)]}
    $x = {[v,\{u_i\}_{i=1,\ldots,|W|}]}$
 \caption{Sentence representation}
 \label{algo:repr}
\end{algorithm}

\subsection{Architectural Details}
\paragraph{Convolutional Layer.} A 1D convolutional layer with filters of multiple sizes \{2, 3, 4\} is applied to the sequence of word embeddings. The filters are used to learn useful translation-invariant representations of the sequential input data. A global max-pooling is then applied across the sequence for each filter output.

\paragraph{Concat Layer.}We apply the concatenation layer to the output of the global max-pooling and the output of \texttt{VADER}. 
\paragraph{Activation functions.}
The activation function used between layers is ReLU~\cite{Hinton_rectifiedlinear} except for the out layer where tanh is used to map the output into [-1, 1] range.
\paragraph{Regularization.}
Dropout~\cite{srivastava2014dropout} was used to avoid over-fitting to the training data: it prevents the co-adaptation of the neurones and it also provides an inexpensive way to average an exponential number of networks.
In addition, we averaged the output of multiple networks with the same architecture but trained independently with different random seeds in order to reduce noise.
\paragraph{Loss function.}
The loss function used is the cosine distance between the predicted scores and the gold standard for each batch. Even though stochastic optimization methods like Adam~\cite{KingmaB14} are usually applied to loss functions that are written as a sum of per-sample loss, which is not the case for the cosine, it converges to an acceptable solution.
The loss can be written as :
\begin{equation}\label{eq:loss}
{\rm Loss} = \displaystyle\sum_{\rm B \in Batches}1 - {\rm cos}(\hat{\textbf{V}}_B, \textbf{V}_B),
\end{equation}
where $\hat{\textbf{V}}_B$ and $\textbf{V}_B$ are the predicted and true sentiment scores for batch $B$, respectively.

The algorithm for training/testing our model is reported in Algorithm~\ref{algo:train}.



\begin{algorithm}
\SetAlgoLined
    \SetKwInOut{Input}{Input}
    \SetKwInOut{Output}{Output}
    \SetKwInOut{Parameter}{Parameters}
    \Input{A set of training instances $S$, with ground-truth scores $y$, and the set of test sentences $S_o$}
    \Output{A set of trained models $M$, and the predictions $y_o$ for the test set $S_o$}
    \Parameter{The number $N$ of models to train}
    ${\rm preprocess}(X)$ \tcp{see sec 3.1}
    \ForEach{$s_i$ in $S$}{$X_i = {\rm sentence\_representation}(s_i$) \tcp{see Alg.~\ref{algo:repr}}}
    \ForEach{$n \in N$}{$M_n = \min {\rm Loss}(X)$ \tcp{see Eq.~\ref{eq:loss}}}
    \ForEach{$n \in N$}{$y_n = {\rm evaluate}(X_o, M_n)$}
    $y_o(u) = \frac{1}{N} \sum_n^N y_n(u)$
 \caption{Training/Testing algorithm. To build our model, we set N=10.}
 \label{algo:train}
\end{algorithm}


 
 	 
\section{Results}
In this section, we report the results obtained by our model according to challenge official evaluation metric, which is based cosine-similarity and described in~\cite{ghosh2015semeval}. Results are reported for three diverse configurations: (i) the full system; (ii) the system without using word embeddings (\emph{i.e.} \texttt{Glove} and \texttt{DepecheMood}); and (iii) the system without using pre-processing.
In Table~\ref{tab:results-cv} we show model's performances on the challenge training data, in a 5-fold cross-validation setting.

\begin{table}[h!]
    \centering
    \begin{tabular}{c|c}
         \textbf{Algorithm}& \textbf{mean$\pm$std} \\ 
         \hline
         Full& 0.701 $\pm$0.023\\ 
         No embeddings& 0.586 $\pm$0.017\\ 
         No pre-processing& 0.648 $\pm$0.022\\ 
         \hline
    \end{tabular}
    \caption{Cross-validation results}
    \label{tab:results-cv}
\end{table}

Further, the final performances obtained with our approach on the challenge test set are reported in Table~\ref{tab:results-test}. Consistently with the cross-validation performances shown earlier, we observe the beneficial impact of word-representations and basic pre-processing.

\begin{table}[h!]
    \centering
    \begin{tabular}{c|c}
         \textbf{Algorithm}& \textbf{Test scores} \\ 
         \hline
         Full& 0.745\\ 
         No embeddings& 0.660\\ 
         No pre-processing& 0.678\\ 
         \hline
    \end{tabular}
    \caption{Final results}
    \label{tab:results-test}
\end{table}

\section{Conclusions}
In this paper, we presented the network architecture used for the Fortia-FBK submission to the Semeval-2017 Task 5, Subtask 2 challenge, with the goal of predicting positive (bullish) or negative (bearish) attitude towards a target brand from financial news headlines. The proposed system ranked 1st in such challenge. 

Our approach is based on 1d convolutions and uses fine-tuning of unsupervised word representations and a rule based sentiment model in its inputs.
We showed that the use of pre-computed word representations allows to reduce over-fitting and to achieve significantly better generalization, while some basic pre-processing was needed to further improve the performance.

\bibliography{acl2017}
\bibliographystyle{acl_natbib}
\end{document}